# A Stylistic Analysis of Honest Deception: The Case of 'Seinfeld TV Series' Sitcom

A paper submitted to the department of English in partial fulfillment of the requirements for the degree of License in English (Major: Linguistics) Academic year: 2019-2020

by
**Mohcine EL BAROUDI**
Hassan II University, Casablanca, Morocco
mohcine.elbaroudi-etu@etu.univh2c.ma
mohcine.elbaroudi@gmail.com

Supervisor
**Moulay Sadik Maliki**
Hassan II University, Casablanca, Morocco
sadikmaliki.moulay@univh2c.ma

## Abstract

Language is a powerful tool if used in the correct manner. It is the major mode of communication, and using the correct choice of words and styles can serve to have a long-lasting impact. Stylistics is the study of the use of various language styles in communication to pass a message with a bigger impact or to communicate indirectly. Stylistic analysis, therefore, is the study of the use of linguistic styles in texts to determine how a style has been used, what is communicated and how it is communicated. Honest deception is the use of a choice of words to imply something different from the literal meaning. A person listening or reading a text where honest deception has been used and with a literal understanding may completely miss out on the point. This is because the issue of honesty and falsehood arises. However, it would be better to understand that honest deception is used with the intention of having a lasting impact rather than to deceive the readers, viewers or listeners. The major styles used in honest deception are hyperboles, litotes, irony and sarcasm. The Seinfeld Sitcom TV series was a situational TV comedy show aired from 1990 to 1998. the show attempts to bring to the understanding the daily life of a comedian and how comedian views life experiences and convert them into hilarious jokes. It also shows Jerry's struggle with getting the right partner from the many women who come into his life. Reflecting on honest deception in the Seinfeld sitcom TV series, this paper is going to investigate how honest deception has been used in the series, why it has been used and what is being communicated. The study is going to use a recapitulative form to give a better analysis and grouping of the different styles used in honest deception throughout the series.

**Stylistics-** The study of the distinctive styles found in particular literary genres and in the works of individual writers.
**Honest deception**- this is the use of words that seem to be deceitful but with an intention to pass a message in a hidden way.
**Hyperbole-** Exaggerated statements or claims not meant to be taken literally. They are often understood as overstatements.
**Litotes-** Ironic understatement in which an affirmative is expressed by the negative of its contrary.
**Irony-** The expression of one's meaning by using language that normally signifies the opposite, typically for humorous or emphatic effect.
**Sarcasm-** The use of irony to mock or criticize.

# Introduction

Language is one of the most powerful tools in human associations due to the lasting effect on the recipients. There are many ways of using language to pass a specific type of message. The way a person uses language is determined by the choice of words, annotated expressions, and the use of figurative speech. There are numerous styles in which a person can strategically use language to indirectly pass a message that would otherwise fail to impact or cause conflicts if passed directly. Language is dynamic and, therefore, can be easily altered to fit varied situations.

Stylistics, derived from the word style, is used to define the various styles in which words and language are used in literature. As a study, it aims at describing the use made up by language. Stylistics is majorly studied to acquire skills of exploring and understanding the linguistic features of texts. Stylistics can also be understood as the link between linguistics and literary criticisms. Stylistics, therefore, serves to join literature and literary criticism.

Stylistic analysis is the analysis of various language styles to extract the meaning or message contained in texts. Stylistic analysis requires critical analysis of texts to recognize the various styles used. Once a person identifies the styles, there arises the need for a good understanding of the styles in order to derive the meaning or the interpretation of the hidden meaning. The skill and understanding are mostly needed when dealing with literary works or poems. This is because poets are very unlikely to pass a message in a factual manner. Unlike historians, they pass their message or information in an indirect and hidden way.

Honest deception is a situation whereby the true nature of things is indirectly described using exaggerating language styles. The main styles used in honest deception are hyperbole, litotes, irony, and sarcasm (Alabi, 2007). Hyperbole styles involve the overstatement in language to send a strong and heavy emphasis on something. An example is describing a person who has overstayed in a place as having taken centuries in the place. With human life limited to less than two centuries, saying a person has taken centuries is an overstatement and exaggeration in the actual sense. However, the statement best fits to be used to indicate that a person has taken a long time to do something.

Litotes, on the other side, can be defined as an understatement. Litotes and hyperbole share close similarities in their structural styling. However, while a hyperbole can be described as an overstatement, litotes are more of an understatement (Alharthi, 2016). Additionally, hyperbole can be seenas optimistic, while litotes are pessimistic. Litotes use the opposite of a situation to describe it. In the case of a rich man, litotes application will describe the man as 'he is no a pauper.

The irony is the use of words in a specific way with an intention to pass a message that is conceived differently from the literal meaning of the chosen words. There are three types of irony, verbal, situational, and mention-tone irony. Verbal irony involves the use of a choice of words to express something that is different from the literal meaning. The situational irony, on the other side, means the actions whose outcomes are different from the expected ones. Dramatic irony involves a situation whereby the audience knows a truth that a character in a movie or literature work is not aware of.

The Seinfeld Sitcom TV series is a show about Jerry and his comedy career. Coming with a new style and dimension that was different from the regular stand-up comedy, the TV series gives a glimpse of Jerry's life. The show further gives a deep insight into the day to day life and the relationship between the life experiences and the type of comedy that a specific comedian focuses on. Additionally, the show details Jerry’s struggle with getting the best woman to settle with. This gives a highlight on the normal struggle of young men at his age in finding the suitable woman and the underlying values that a man and woman want in a partner.

In the TV series, we get to understand how daily social interactions impact jerry's career. Most of the jokes he makes to the audience are closely tied to the activities and interactions that occurred to him recently. Beginning with the interactions with Laura, who he wants to date before realizing she is engaged, Jerry gives a comedy session about the difficulty men have in picking signals and even expressing themselves to women. The same is seen when thieves break into his home and steal some items from his house while he is away. In the following comedy session at the comedy club, he satirically criticizes having to call the police who seem to have no clue of what happens around.

The TV series further uses different language styles to explain the day to day life in the real world. It uses the styles to pass a message in a way variant to the literal way and to criticize some ills in the society. Jerry takes on various issues, such as misunderstanding among parents and their young ones. He also criticizes the inefficiency of some government authorities. Furthermore, he outlines young people's daily struggles with no regular cash flow and just a normal hustle. He further uses the different styles to highlight the comedy industry, which, according to him, is selling nothing.

## Literature Review

A research paper on the stylistic analysis of Advertising discourse, a case study of Dangote cement in Cameroon by Fomuking, outlines the use of language styles to persuade. The paper analyses the use of different language styles in the advertising strategy to leave a lasting impact on the advertisement's viewers. The analysis is carried out through an analysis of what is being communicated, how it is communicated, and the interpretation.

The study on stylistics analysis of advertising discourse studies the various ways that styles can be used to promote a persuasive motion. The study details how various styles have been usedin listing, prioritizing, assuming, naming, and describing (Fomukong, 2016). The paper contends that in an advertisement context, the advertiser uses the language to build trust and a good relationship with the audience who are the consumers. The producer, who might be the producer, uses grammatical patterns and lexical items to develop social interactions. This is because the consumer cannot be coerced to consume based on the persuasion.

Leech, in his book on honest deception, unveils the mask of irony. He contends that irony is a style that requires a two-side audience; one in the know and one in the dark. The dark side is naïve to the extent that they are willing to take statements at face value (Leech, 2014). As per the study, irony has two intentions. The first intention aims at concealing whatever is being said. The second motive is that whatever is being concealed is meant to be unveiled.

In his book Seinfeld and the comedy vision, Kaufman explores the use of irony in the TV series. He contends that Jerry uses irony to portray the situation in the real world. According to Kaufman, Jerry's irony portrays the hypocrisy in people, lying all the time and not regretting it. Ironically, the characters in the TV series lie, but readily admit to lying. Those who lie do not go far with life before getting caught. This is seen in various instances, such as when George and Jerry pretended to be Murphy and O'Brien without knowing who those really were. Jerry also lies that he is tutoring his nephew not to meet with Joel only for him to visit him and find out the lie.

# Method

The research paper aims at performing a stylistic analysis of the Seinfeld Sitcom TV series. The paper aims at finding out how the different styles of honest deception have been used in the Seinfeld TV series and their respective meaning. The paper identifies how the different styles have been used to communicate, what is being communicated, and the interpretation. The paper focuses on hyperbole, litotes, and irony used in honest deception.

The study is carried out through the Textual Conceptual Functions (TFCs)(Jeffries, 2016). According to the TFCs, a stylistic determination is done by the use of the evidence from the text and the use of stylistic models for the analysis. The judgments then give the perspective at which the text functions. According to Jeffries, a description of the language used in a text is linked to various interpretations. There are three meanings that may be attached to the language of a text; the linguistic, ideational, and interpersonal.

The linguistic meaning involves the structural and semantic and de-contextual that examines the level of languages used, such as phonology, morphology, syntax, and discourse. The linguistic meaning gives the primary meaning to all language use (Li & Shi, 2015). The ideational meaning, also known as the textual-conceptual meaning, is the co-textual effects of linguistic choices made by language users. In examining the ideational meaning, the analyst considers the linguistic environment in which the word has been used in the text. This then gives the various meanings usable in analyzing the text and the determination of the various ways in which texts represent the world. The interpersonal meaning includes the practical meaning focusing on the contextual intention of language use on other people. At this level, the analyses are on the way interlocutors interact and the ways the producers seek to achieve their intentions by using language in speech acts, and implicatures.

# Results and Discussion

The results from the Seinfeld Sitcom TV series' structural analysis portray a vast usage of language styles to pass messages in the various conversations. Honest deception in the TV series is evident throughout the series. Honest deception in the TV series is passed through various instances of hyperbole, litotes, and irony.

## Hyperbole

Hyperbole is one of the styles used in the Seinfeld series. The style is used to stress some statements in which a direct statement would not effectively drive the weight of the information needed. Leech, in his article on honest deception, a guide on linguistic poetry, describes hyperbole as a style used on the personal opinion of a person to another. He contends that the exaggeration made is only unless we are unable to get into the person's personal life to whom the compliments are made on to confirm the truths. Hyperbole are exaggerations made for the sake of emphasis and not foe deception as could otherwise be literary understood.

The first incident of hyperbole is seen at the beginning of the season one episode one. Immediately after the first comedy session at the comedy club, Jerry uses hyperbole in describing George’s buttoning of his shirts. “Seems to me, that button is in the worst possible spot,” he comments on the tightness of the button. “The second button literally makes or breaks the shirt, look at it: it's too high! It's in no-man's-land.” As much as the button is placed too high in the shirt, saying that it is in no man’s land is an extreme exaggeration. However, the statement is the best suited in describing the position of the button in an abnormal position.

In the incident where Kramer is commenting about the tapped game, Jerry makes fun of him for staying indoors for too long. "Yeah, you almost went to the game. You haven't been out of the building in ten years!" It is impossible for a person of good health and without legal restrictions such as a house arrest to stay indoors for ten years. However, Jerry uses the style to make a strong statement of the length of time Kramer has been in the house. Jerry’s overstatement serves to emphasize the Kramer's behavior of always staying at home and the little connection with the outside world. It can be viewed as a strong persuasion to Kramer to try to get out of the house and experience life in the outside world. This is backed by Jerry's opinion when he says that everyone looks forward to going out and, once out, starts thinking of going back home to sleep.

Hyperbole is also seen in the comedy session when Jerry talks about girls and cotton-balls. He explains his struggle with having to buy cotton-balls whenever a girl is paying him a visit. Despite never being in a situation when he needs to use a ball or be aware of how to use one, he still has to find a way to have cotton-balls in his washroom during such visits. However, Jerry exaggerates the number of balls that a girl needs in a single visit. He says,”they need thousands of cotton balls.” He further says one moment there are many of them in a huge bag, and the next moment they are all used up. The number of cotton-balls he states is obviously exaggerated. However, he decides on such a number to emphasize the extreme demand for cotton-balls that women need and use for make-up and other reasons. In the same session, he also exaggerates how girls use the few balls that they decide actually to use. He describes the used cotton-balls in the waste-bin as have undergone intense torture and interrogation by the ladies.

When the doctor comes to Jerry’s home to check up on his father’s back pains, Jerry’s mother makes fun of his son’s old couch. "One day somebody's going to sleep on that thing, and we'll get sued. I hope this doctor knows what he's doing." By these words, he describes the state of the coach whose cushion has worn off to an extent the supporting wood presses on whoever sleeps on it. Despite giving a person some pains, resulting in some legal case is an overstatement of the situation.

**Litotes**

Litotes are another linguistic style used in honest deception. Litotes are statements that take the opposite form of the intended literal description (Carmen & Malcolm, 2013). The use of litotes is seen when Jerry's parents are helping him plan how to meet with a lady he likes. The plan is to pretend that Jerry and his friend are meeting a friend in the same building the lady is coming from. When the parents lay out the idea to Jerry, he comments, "Y'know what? This is *not* that *bad* an idea”. Jerry likes the plan and thinks it is a good idea but uses an understatement to mean that he appreciates and will use the plan.

In the discussion about investing in stocks, Georg lays out a good investment plan for Jerry. Jerry has always had a negative attitude toward investment. He even openly criticizes the advice on letting money work for you rather than you working for money. He opts to work and let the money relax. About the investment plan, he says that it is not that bad. George feels that Jerry is understating the profits and income that the investment may bring. He tells Jerry that it is, in fact, a fabulous idea.

**Irony**

Irony is also widely used in the Sitcom TV series to portray Honest Deception. The three types of irony, verbal, situational, and dramatic irony, have been widely used in the TV series. Verbal irony is the use of a figure of speech in one way but to mean the other. Verbal irony is meant to be understood in a way that is varied to the literal meaning if the statement. Situational irony, on the other side, is used in a situational manner or incidents. Situational irony involves incidents that have an outcome that is contrary to the peoples' expectations.

Jerry uses immense irony to visualize various truths of life. He craftily discusses hypocrisy, misunderstanding, and the various daily challenges that a normal person undergoes. One of the major types of irony that Jerry uses is situational irony. He explains how people view the lifestyle of going out. When in the house, people get bored and start thinking about going out and having fun. They do this to break the monotony and refresh their minds. Jerry details the efforts that people put into looking gorgeous for the outs. People take their time to pick the finest clothes, arrange for transport, call their friends and even make reservations. Ironically, however, once people get to the destinations, they start looking at their time and now want to go back to the same boring environment they were escaping. Long hours of sleep forced them out, and now they want to go back to the same sleep again.

Another irony comes from Jerry's description of his career. While others chose different paths, he chose to be a comedian. In one of the comedy sessions, he says if there is someone who has a life with no point is him and his comedy career. When he and his father are arguing about who is going to pay for gas at the gas station, he is agitated by the feeling that his father thinks that he does nothing. It is, however, ironic that he always sees his career as selling nothing when it is his self-chosen career path. Additionally, comedy is what keeps his life moving and even gives him his daily bread. It takes his close friends' effort to convince him that the comedy career is not as bad as he thinks.

After George and Marlene's break-up, Jerry volunteers to get back the books from Marlene's place because George felt it was awkward going back to Marlene's place. As a friend,it is unexpected for Jerry to have an intimate relationship with Marlene. Despite George breaking up with Marlene, his tone tells that he has not completely got Marlene off his thoughts. Jerry, however, goes ahead and almost falls in love with Marlene. Taking into consideration that Jerry advised George on how to break up with Marlene, it is ironic that he goes ahead and gets into a brief relationship with her. Additionally, Jerry states that he wants the relationship with Marlene to end. However, it is ironic that he is not happy when Marlene tells him that she does not want to be in the relationship anymore. Dramatic irony is also seen in the situation where Kramer and Jerry are aware that Jerry and Marlene are already getting intimate, but George does not. George thinks the moment Jerry confesses his affection is the moment he seeks permission to get into a relationship with the girlfriend.

In the race situation, when Jerry takes a head start before everyone, it is conceived that he is fast in track events. The teachers and fellow students believe Jerry has a talent apart from Duncan, who suspects something. Jerry is aware that he is not the fastest and decides that he will never participate in any other race. Everyone is made to believe that Jerry despises racing for some other reason, but only Jerry himself knows he is maintaining the legacy. Jerry makes it aware to everyone that the issue of ever racing again is out of the question. However, he accepts to participate in a race that he plans to give an excuse to avoid.

In the meeting with Duncan, George and Jerry plan to trick Duncan into believing that the famous race was logistically worn. There are several openings that could easily have made it obvious to anyone that the two were acting not to know each other. Duncan is, however, unable to realize it. Being four former classmates, anyone could wonder why Jerry was only interested in knowing more about George than any other person. Secondly, it is awkward for George to suddenly appear into a meeting, which was only meant for Jerry, Lois, and Duncan. Thirdly, it is not normal for a person to meet former classmates and leave suddenly before they have known much about each other. Further, it is weird for a full-grown man to so accurately remember a race that occurred years ago during school days. Ironically, Duncan is unable to see all of this and realize that the meeting is a hatched plan to convince him.

Jerry further uses situational irony to criticize robbery incidents, and the police department's unreliability to catch the criminals. After the robbery incident, he details how the police come and hand him forms to fill, which are to be used in the investigation. Despite refraining from political biasness, he criticizes the robbery victim having to fill the forms when it is clear that it is almost impossible for the current police department to carry out an effective investigation. Jerry contends that if the strategy is to work, the thief should also be handed a form to fill to ease the investigation process. This is ironic in that a suspect will not have to fill the form as he will be already captured and, therefore, no need for an investigation.

Verbal irony is seen in several incidences in the series. The first incident is in the first episode of the first season when George is describing Jerry's relationship with girls when Laura calls to visit him. He says, Jerry, "God bless, you devil." Taken literary, God's holy blessings to a devil is a deception and an impossibility by itself. However, George uses the word Devil to ironically describe Jerry, who seems to have a better spot for girls. Additionally, the word can also be seen as an indication of Jerry's ability to interact socially with many girls.

Verbal irony is also seen when Jerry is filling out the police forms when the investigation team comes to his house to investigate the robbery incident. He criticizes the ironic situation whereby police always fill have victims fill out forms but never come up with the results of investigations such as arrests of the perpetrators or the returning of the stolen goods. Verbal irony is evident in their conversation.

POLICEMAN: I see...Well, mister Seinfeld uh, we'll look into
it and uh, we'll let you know if we uh, you know if we find
anything.
JERRY: You ever find anything?
POLICEMAN: No. [hands Jerry his copy of the report]
JERRY: Well, thanks anyway.

POLICEMAN: You bet.

When jerry asks the police if they ever find anything, they will contact him. He stylistically expresses his frustration with the police department on the failure of ever coming up with a conclusive investigation. The question further shows his contentment in the fact that the police will not find the perpetrator. That he is already aware that there is very little hope of getting the police department's assistance in getting back his stolen items. His conclusive opinion is seen when he sarcastically criticizes his friends telling him to call the police when in reality, he has had numerous robbery incidents with none of the police investigation being fruitful.

Another incident where Jerry uses verbal irony is in the first season first episode when Kramer comes to borrow his meat. Jerry is in the middle of a game and has all concentration on the game. He did not want anyone to give him a clue of how the game ended but wants to know it from the tapped game. He stays late into the night with the expectation that no one will interrupt him at that hour of the night. Just then, Kramer bursts in commenting about the game, which irritates Jerry. He then announces the purpose of the visit, which is to borrow meat. Jerry hungrily tells him that he does not have any neat, and if Kramer needs it so much, he could as well as go out and hunt for his own meat. The irony in the conversation arises in how easily Jerry satirizes Kramer, who leaves alone going hunting, looks too clumsy to perform heavy duties, and rarely even gets out of his house to get supplies. Hunting is far away from the thought.

The incident also brings up another irony when Kramer wants to help Jerry convince Laura to keep on the plan of coming over on the weekend. Kramer acts like he knows his ways with women. However, judging by the situation, he almost has no idea of how to treat women rightfully. As a matter of fact, he is way too older than Jerry but is still not married. It is ironic that he pretends to be able to convince women, yet he himself has not been able to convince any woman to become his girlfriend or wife. This fact is ascertained by how he looks astonished when Jerry buys an additional bed for the visiting Laura. He keeps asking Jerry why he is giving the lady options.

Jerry has never had any interest in indulging in the stock market business. He believes the stock trading is a gamble where you put your money and are not sure of returns. As a matter of fact, he believes once you put your money in stocks, they begin decreasing until they are done. When George introduces him to the trading idea, he sees it as a scam and doubts all individuals involved in the trading activities. Jerry refutes the idea of money working for a person and decides to work and let his money relax. Nonetheless, he gives the stocks trading idea a try but with a very pessimistic attitude. When the prices go low, he confirms his fears and begins counting his losses. He ends up calling up the sell to avoid further losses. He intentionally forgets about the idea only to realize that the prices rose, and people made profits after withdrawing his money. This leads to the topic of his following session, where he sarcastically refutes the trading business.

**Sarcasm**

Another honest deception style widely used in the series is sarcasm. Sarcasm is seen in most of Jerry's comedy sessions. When he is planning on meeting Laura, he talks about how it is a challenge for men and women at the marrying age to find their perfect match. He satirically tells the story of a man honking at a lady expecting the lady to know that he is interested in her. It is, however, weird for a person to interpret an irritating honking as a sign of interest in someone. He also tells how he faces challenges picking up the signal. Jerry sarcastically details how the ladies, on the other side, struggle to find a man. Jerry is bewildered by how ladies will go as far as reading large books to know where and how to get men when he and other men are everywhere.

Jerry further employs sarcasm at the comedy club. He says after being robbed eighteen times; he does not see the need for involving the police in the investigations. He explains his disappointment with his friends, who tell him to call the police. Sarcastically, he states that it is impossible to capture the crook if they do not give them a form to fill.

JERRY: If there's a serial killer lose in your neighborhood; it seems like the safest thing is to be the neighbor and 20% these women who always fall in love with the serial killer. They write to him in prison. Here's a woman that's hard to disappoint. I guess she's only upset when she finds out he's stopped killing people and she goes: "You know sometimes I feel like I don't even know who you are anymore". In the comedy session, Jerry uses sarcasm and irony to detail the plight of serial killers and their prey. Serial killers are highly manipulative. They get the attention and the trust of their victims so as to get the best chance to make the kill. It is, however ironical and hilarious how Jerry relates the neighbors of the serial killers and the heinous acts done by the evil people. He says that they are the safest because they are to give a sermon on how good and humbled the man was. Jerry contends that many women are also likely to fall in love with the serial killer. This is ironical since killers are dangerous people and ladies' value their safety first. It is also awkward that the ladies keep contacting the killers even when they have already been jailed.

Sarcasm is also seen in Jerry's joke about the detergent advertisement. The advertisement purports that it is effective in stain removal and can even remove blood stains. The advert astonishes Jerry, and he does not see why the advertisers would use blood to promote their products. It is sarcastic that the advert only looks at the cleaning of the shirt rather than the source of the blood. "I mean, I, come on, you got a T-shirt with blood-stains all over it, maybe laundry isn't your biggest problem right now." According to Jerry, the advert is contextually wrong. He believes the owner of the shirt will be more concerned about the injury that led to the bleeding rather than getting the blood from the shirt.

# Conclusion

In conclusion, stylistics are linguistic styles that can be used to pass a message or information effectively. Stylistics is may also be used in indirect communication. Language being a powerful tool in human life, the choice of words, expressions, and figurative speech can be altered to determine the use of language and to suit different uses and situations. This can be seen in the historical use of different language styles to criticize or ridicule authority and undesired behaviors in society. The use of words to criticize comes from the joint that styles form between linguistics and literary criticisms.

Stylistics analysis involves the analysis of various styles used in language to derive the desired meaning or theme of given texts. It involves identifying the various ways in which authors of literary works or poets convey their message, i.e., through an indirect and hidden pattern. The stylistic analysis calls for skill and a good understanding of the various uses of language styles in audio or written works. The stylistic analysis involves identifying the styles and the derivation of the message or the meaning of literature work.

Honest deceptionis defined as the stylistic communication where the choice of words used is different to how the situation could be explained literary. The question of honesty and falsehood arises from the words (Adams, Matu&Oketch, 2014). In the literal meaning, the language used could be a total lie.The commonly used styles in honest deception are hyperbole, litotes, and irony. Sarcasm may also be employed. While hyperbole is figurative statements that overstate a situation, litotes are understood as understatements. Irony, on the other side, involves a choice of words that mean the opposite of the literary meaning.

This research paper seeks to perform stylistic analysis on the Seinfeld Sitcom TV series. It focuses on the honest deception in the TV series by identifying the use of hyperbole, litotes, and irony. Analysis of the paper involves identifying what is being communicated, how it is being communicated, and the interpretation of the choice of words used in the linguistic styles. The study employs

There is a wide usage of hyperbole throughout the series. Jerry, a comedian, is fond of using hyperbole in his normal communication and comedy session to drive a lasting and weighty impact of his craft. Hyperboles seem to be one of his strategies of effectively communicating with everyone around him. This is seen throughout the series beginning from the first episode when he states that George's shirt button is in no man's land. He also comments on Kramer's overstaying indoors by stating that he has not gone out for over ten years. When he is leaving, and Elaine decides to spend the weekend in his house, Jerry comments about the state of his house saying that the expiry date of things in the refrigerator will have to be subtracted two days.

Irony is also employed throughout the TV series. This is seen in various scenarios, for instance, when he decides to try out stock trading, a thing he had not liked from the word start. He loses some money, and others accrue profits once he withdraws from the trade. Irony is also seen in Jerry's explanation of the people's opinion of going out. He says that people long to go out only to long to go back to the house once they get out. It is also ironic that Jerry conceives signing a comedy contract as selling nothing when comedy plays a central role in his life in meeting his bills and associating with people. Also, Jerry advises George on how to directly approach Marlene and end the relationship. It is ironic that he gets together with his best friend's ex-girlfriend, whom he is unable to follow the same advice he offered George and approach her to end the undesired relationship. It is also ironic that Jerry, with his knowledge of relationships, is unable to realize that Marlene is only trying to fill up the gap left by George by getting into a relationship with him.

Irony is also seen in Duncan's inability to realize that Jerry has coached George into the meeting to cement the belief that he won the race legitimately. The audience is aware of the fact that Jerry had a head start which the coach failed to see and helped him win the race, and that he decided never to race again only to keep the legacy. Irony is also seen in the police having people fill up forms for investigation purposes when, in reality, none of the previous investigations had reliable results. It is also ironic that Jerry advises people on some personal principles and promises which he breaks. He dislikes stocks trading but still invests, leading to losses. He also promises to never take part in any race but ends up agreeing to race with Duncan to affirm his school days' victory.

Jerry further employs sarcasm in most of his jokes. He gives an ordeal of the lack of coincidence between searching ladies and gentlemen. While men are unable to pick signals, ladies are busy reading about where to get men who, according to Jerry, are everywhere. He also uses sarcasm in the event of being advised to call the police to assist him in recovering his stolen items when none of the eighteen previous investigations helped. Jerry uses litotes mostly in belittling ideas given. When given very life-changing ideas, he replies that the ideas are not that bad when, in fact, the ideas could very much help him in his life.

Sarcasm is also seen in the joke about the advert. Jerry criticizes the advertisement that promotes the stain removal power of the detergent being advertised using blood. According to Jerry, a normal person will be interested in preventing life loss of life from the bleeding rather than having the shirt clean. In the issue of the serial killer, Jerry seems to be taken aback by the manipulative power of the serial killers. They seem to attract the complete love of women who follow them blindly to their deaths. Even when the killers are jailed, the girlfriends still make efforts to reach out to the killers.

# APPENDICES

| Datum number | Code | Dialogue | Hyperbole | Litotes | Irony | Sarcasm | Explanation |
|---|---|---|---|---|---|---|---|
| 1 | Good news bad news (00:19:35) | JERRY: [a little irritated] Meat? I don't, I don't know, go...hunt! | | | ✓ | ✓ | It's ironical that Jerry tells, Kramer, who never leaves his house to go out and hunt. |
| 2 | (00:00:32-48) | There you're staring around, whatta you do? You go: "We gotta be getting back". Once you're out, you wanna get back! You wanna go to sleep, | | | ✓ | ✓ | Jerry analyses how people go out and get bored immediately they are out and then wish they were back at their houses |
| 3 | (00:1:43) | The second button literally makes or breaks the shirt, look at it: it's too high! It's in no-man's-land, | ✓ | | | | Jerry magnifies the position and tightness of George's shirt buttons. |
| 4 | (00:18:56) | JERRY: [cynical] Yeah you almost went to the game. You haven't been out of the building in ten years! | | | ✓ | ✓ | Jerry makes an overstatement of the longevity of time that Kramer has stayed indoors. |
| 5 | | KRAMER: Oh, I can be very persuasive. Do you know that I was almost... a lawyer. | | | ✓ | ✓ | It is ironical that Kramer acts like women's experts when prior sentiments by Jerry prove he has little connection with the outside world |
| 6 | (00:06:17) | GEORGE: So, ya know, she calls and says she wants to go out with you tomorrow night? God bless! Devil you! | ✓ | | ✓ | | George magnifies Jerry's relationship with ladies |
| 7 | Ex-Girlfriend (00:10:45) | I mean, I, come on, you got a T-shirt with blood-stains all over it, maybe laundry isn't your biggest problem right now | | | ✓ | ✓ | The advert creates more worries than the impact it is driven. A person has to be injured for the detergent to have a purpose. |
| 8 | Good bad | Women need them and they | ✓ | | | | |

| | | | | | | | | |
|---|---|---|---|---|---|---|---|---|
| | news (00:14:12) | don't need one or two, they need thousands of them, | | | | ✓ | |
| 9 | The robbery (00:23:00) | JERRY: There's more, the refrigerator. [Opens it] Deduct a minimum of two days of all expiration dates. | ✓ | | | | Jerry exaggerates the faultiness of his refrigerator |
| 10 | The robbery | POLICEMAN: I see...Well, mister Seinfeld uh, we'll look into it and uh, we'll let you know if we uh, you know, if we find anything.<br>JERRY: You ever find anything? | | | ✓ | ✓ | Jerry makes a joke about the unreliability of the police to make conclusive investigations. |
| 11 | The robbery | Now...unless they give the crook his copy, I don't really think we're gonna crack this case, do you? | | | ✓ | ✓ | It is almost impossible for the police to catch the crook. |
| 12 | The ex-girlfriend | JERRY: I feel terrible. (Kramer smiles) I mean, I've seen her a couple of times since then, and I know I can't go any further, but. I've just got this like, | | | ✓ | | Jerry is unable to follow his own advice and end her relationship with Marlene. |
| 13 | The race | GEORGE: But I digress. Let me see, now. You were standing at one end of the line, and I was right next to you. And I remember we were even for like, the first five yards and then, BOOM,...You were gone. | | | | ✓ | It is ironical that Duncan does not realize that the meeting is all well planned. |
| 14 | The race (00:19:29) | Duncan wants to get together with her and me for lunch tomorrow. He obviously wants me to admit I got a head start. | | | ✓ | ✓ | Jerry later accepts participating in the race, attempts another race which he wins again. |
| 15 | The race | GEORGE: He wants to meet you?I'll tell you what. I'll show up. He doesn't know we're friends. I'll pretend I haven't seen you since High School. I'll back up the story.<br>JERRY: That's not bad. | | | ✓ | ✓ | Duncan is unable to realize he is being lied to. |
| | | GEORGE: Not bad? It's gorgeous! | | | | | |
| 16 | The wallet | JERRY: You don't think I make money. That's what you think, isn't it? | | | ✓ | | It is ironical that Jerry still assumes people think his career is a joke; he knows better that it is well paying. |
| 17 | The wallet (00:23:08) | JERRY: We're lucky they are even interested in the idea in the first place. We got a show about nothing. With no story. | | | | ✓ | Here, Jerry seems to be fighting with his personal thoughts comedy is not a job like accounting and doctors. |
| 18 | The smelly car (00:01:18) | Wife: Andrew, why do you have to pick your teeth at the table?<br>Husband: Leave me alone.<br>Jerry: Yeah, I'm wanting to get married *real* soon... | | | | ✓ | It is sarcastic that Jerry says he wants to marry after witnessing a quarrel.<br>In actual sense, he is depicting the differences in the life of single and married people. The sarcasm here is used to show that he better not marry. |
| 19 | The masseuse (00:00:05-14) | JERRY: If there's a serial killer lose in your neighborhood; it seems like the safest thing is to be the neighbor.<br>20% and these women who always fall in love with the serial killer. They write to him in prison. Here's a woman that's hard to disappoint. I guess she's only upset when she finds out he's stopped killing people and she goes: "You know sometimes I feel like I don't even know who you are anymore". | | | ✓ | ✓ | Jerry is astonished at how serial killers are so convincing. It is absurd that their lovers are so in love with them that they get worried when the killer stops their vicious acts. |